\documentclass[letterpaper, 10 pt, conference]{ieeeconf}  

\IEEEoverridecommandlockouts                              

\title{\LARGE \bf
Self-Aware Adaptive Alignment: Enabling Accurate Perception for Intelligent Transportation Systems
}
\usepackage[T1]{fontenc}  
\usepackage{amsmath}  
\usepackage{amssymb}  
\usepackage{graphicx}  
\usepackage{subcaption} 
\usepackage{float}  
\usepackage{array}  
\usepackage{tabularx}  
\usepackage{color}  
\usepackage[dvipsnames]{xcolor} 
\definecolor{c1}{HTML}{52A051}  
\definecolor{c2}{HTML}{E83100}  
\usepackage{cite} 
\usepackage{hyperref}  
\usepackage{cleveref}
\crefname{figure}{fig.}{figures}
\Crefname{figure}{Fig.}{Figures}

\author{Tong Xiang$^{1}$, Hongxia Zhao$^{2}$, Fenghua Zhu$^{3}$, Yuanyuan Chen$^{4}$, Yisheng Lv$^{5}$
\thanks{*This work was supported by Jiangxi Provincial Natural Science Foundation No. 20232ABC03A07 and the Youth Innovation Promotion Association, Chinese Academy of Sciences (No. 2023141). (Corresponding author: Fenghua Zhu).}
\thanks{The authors are with the State Key Laboratory of Multimodal Artificial Intelligence Systems, Institute of Automation, Chinese Academy of Sciences, Beijing 100190, China {\tt\small (e-mail: xiangtong2023@ia.ac.cn, hongxia.zhao@ia.ac.cn, fenghua.zhu@ia.ac.cn, yuanyuan.chen@ia.ac.cn, Yisheng.Lv@ia.ac.cn).}}
}
\usepackage[switch]{lineno}  
\begin{document}

\maketitle
\thispagestyle{empty}
\pagestyle{empty}

\begin{abstract}

Achieving top-notch performance in Intelligent Transportation detection is a critical research area. However, many challenges still need to be addressed when it comes to detecting in a cross-domain scenario. In this paper, we propose a Self-Aware Adaptive Alignment (SA3), by leveraging an efficient alignment mechanism and recognition strategy. Our proposed method employs a specified attention-based alignment module trained on source and target domain datasets to guide the image-level features alignment process, enabling the local-global adaptive alignment between the source domain and target domain. Features from both domains, whose channel importance is re-weighted, are fed into the region proposal network, which facilitates the acquisition of salient region features. Also, we introduce an instance-to-image level alignment module specific to the target domain to adaptively mitigate the domain gap. To evaluate the proposed method, extensive experiments have been conducted on popular cross-domain object detection benchmarks. Experimental results show that SA3 achieves superior results to the previous state-of-the-art methods.


\end{abstract}

\section{Introduction}

Object detection is a challenging problem in computer vision with extensive real-world applications such as intelligent transportation. This task aims to identify and locate the object of interest from the given images or video frames. However, the performance of object detection models often degrades dramatically when applied to a new domain or dataset, which is known as the domain shift problem. 

To address the domain shift problem, recent studies \cite{b1,b2,b3,b4,b5} have proposed various domain adaptation networks. Adversarial learning through domain classifiers \cite{b6} is the initially proposed methods and has greatly improved accuracy in different cross-domain scenarios. Though it can generalize well across domains by aligning the statistical distributions between the source and target domains, solely using adversarial learning on the complex detection task still has strict limitations.

To explore the potential of self-training on the unlabeled target domain for improved detection performance, researchers have shown an increased interest in a novel teacher-student self-training method \cite{b3,b5,b7}. Unbiased Mean Teacher \cite{b3} leverages CycleGAN \cite{b8} to augment the generalisation of teacher-student framework. Also, adaptive Teacher \cite{b5} combines teacher-student model and adversarial learning to suppress the false positive ratio of pseudo labels and update the weights of teacher model with cross-domain student model, which has resulted in a target detection model that is biased towards the target domain but with the style of the source domain. Such a pairing framework still cause false-positive samples, while the large amount of iterative training time may impact the efficiency and real-time performance of the model.

\begin{figure}
    \centering
    \includegraphics[width=1\linewidth]{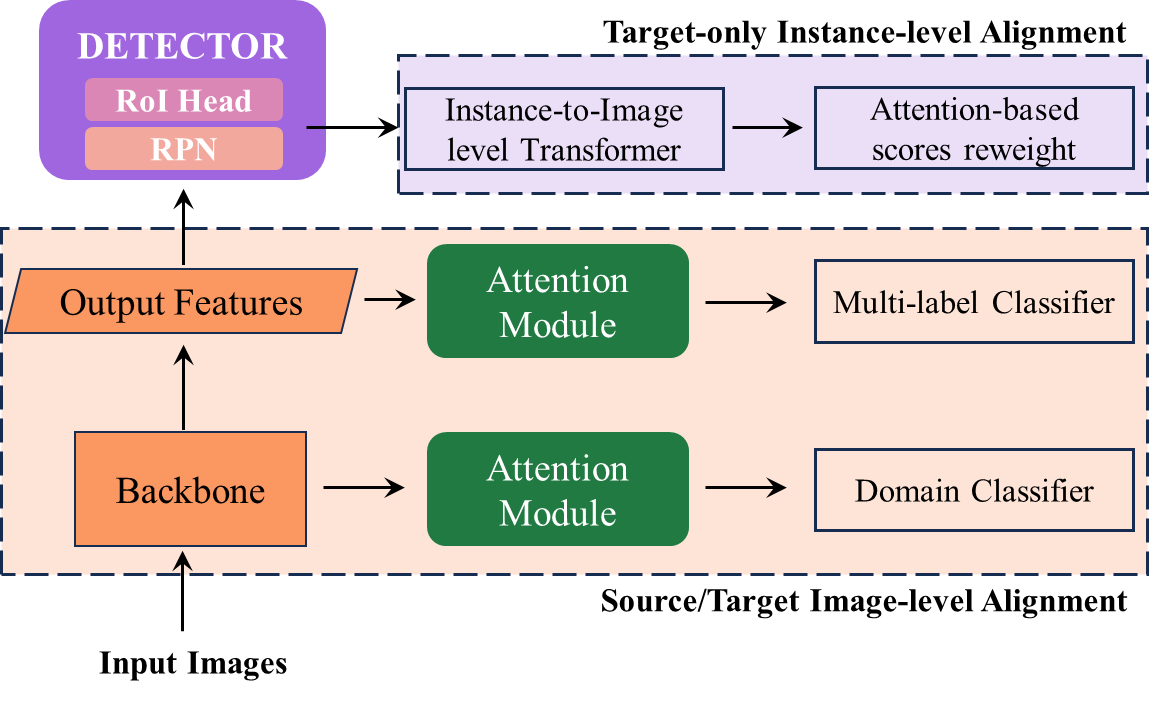}
    \caption{SA3 employs two alignment modules for CDOD (Cross Domain Object Detection).}
    \label{fig:111}
\end{figure}

In this paper, we propose a new network called Attention-based Adaptive Alignment (SA3), the overall framework of this model is shown in \Cref{fig:111}. From the perspective of salient features extraction, we introduce an Attention-based Image-level Alignment Module (AIAM) into the backbone of model, which enhances the representation power of important features and mitigates the discrepancy during the training period, accelerating the process of invariant feature learning and local-global alignment. Furthermore, we accordingly introduce an Instance-to-image level Transformation Module (I2ITM) to assist in the better implementation of the instance-level alignment and mitigate the domain discrepancy between different domains.
The main contributions of this paper are summarized as follows:

\begin{itemize}

\item Two adaptation modules, i.e., Attention-based Image-level Alignment Module (AIAM) and Instance-to-image level Transformation Module (I2ITM), adaptively facilitate image-level global-local alignment and instance-level alignment.
\item Extensive experiments are carried out in three major benchmarks in terms of two typical scenarios, and the results are state-of-the-art, demonstrating the effectiveness of the proposed approach.

\end{itemize}

The rest of the paper is organized as follows: Section II provides a comprehensive review of related research in cross-domain object detection. The proposed SA3 method for adaptive detection of autonomous vehicles is illustrated in Section III. Section IV provides evaluation results and analyses on two cross-domain detection scenarios. The conclusions of this paper are made in Section V.

\section{Related Work}

In advanced driver assistance systems and autonomous vehicles, traffic object detection and recognition play an essential role in the aspect of vehicle perception, which requires solving both recognition and localization. In recent years, object detection networks based on DCNN have achieved good detection performance on a variety of benchmarks.

\subsection{Object Detection}

Among all the works \cite{b9,b10}, the most classical two-stage framework is Faster-RCNN \cite{b11}, which introduces the region proposal network and fuses the region proposal and detection networks, effectively capturing and aligning important local regions, leading to better performance in cross-domain. Also, Faster R-CNN employs explicitly encoding instance-level features through RPN, allowing to focus on relevant instances while ignoring background areas, reducing the amount of unnecessary adaptation required. In this work, we verify our model using Faster-RCNN as the base detection network.

\subsection{Domain Adaptation}

In the world of multiverse, the detection models we obtain from real-world training are usually tough to adapt, which requires us to optimize the model adaptability and implement model generalization as much as possible. Domain Adaptation methods are widely used in many real-world applications, including image-level style shifting and object recognition. Traditional approaches aim to minimize the distance metric between domains, such as Maximum Mean Discrepancy, but with the introduction of Gradient Reverse Layers \cite{b12}, min-max optimization can be constructed through the domain discriminators and feature extractors, this adversarial training is also well-suited for learning domain-invariant features. Some typical networks based on CycleGAN \cite{b8} can achieve diverse transformations between different domains through a cycle-consistent adversarial training process. 

\subsection{Cross-Domain Object Detection}

DA-Faster RCNN \cite{b6} is the first work in cross-domain object detection that designs image-level, target-level discriminators and image-and-instance consistency function to reduce domain gap. Multi-adversarial Faster RCNN \cite{b13} minimizes the domain distribution disparity by aligning domain features and proposal features hierarchically. MTOR \cite{b7} is proposed on top of MT \cite{b14} and considers its object relation by enforcing the region-level, inter-graph, and intra-graph consistency. Furthermore, some works try to address domain adaptation through the style transfer method. UMT \cite{b3} augments the training samples with CycleGAN \cite{b8} and trains the student model with mixed images jointly, effectively mitigating the issues of biased pseudo-labels and class imbalance in object detection tasks. 

\section{Proposed Method}

In this section, we describe the proposed self-aware adaptive alignment methodology and its implementation. Our approach strives to endow the domain adaption model with self-awareness detection by introducing two modules Attention-based Image-level Alignment Module (AIAM) and Instance-to-image Transformation Module(I2ITM), with the ability to adaptively adjust the weights to focus on the important features and facilitate global-to-local alignment.

\begin{figure*}
    \centering
    \includegraphics[width=0.9\linewidth]{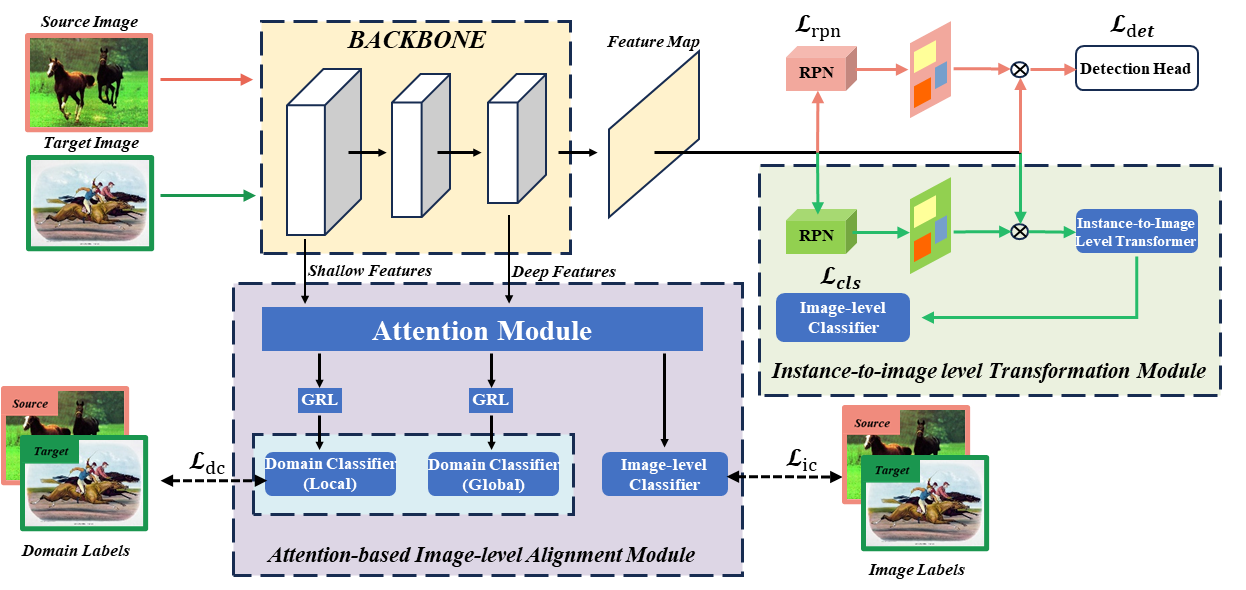}
    \caption{Framework of our Attention-based Adaptive Alignment R-CNN.}
    \label{fig:222}
\end{figure*}

\subsection{Framework Overview} 

The architecture of our proposed SA3, which adopts a two-stage detection framework Faster R-CNN \cite{b11} comprised of backbone, RPN, and detection head, as illustrated in \Cref{fig:222}.
SA3 takes the mixture of source and target images as its input and splits them into two batches annotated with domain labels for domain-specific training apart from the Faster R-CNN series. To this end, we improve two modules based on the baseline structure: attention-based image-level alignment module (AIAM) and instance-to-image transformation module (I2ITM). 

AMAI proceeds along the backbone of the detection pipeline \cite{b11} and implements attention-based global-to-local image-level alignments via an attention adaptor, two domain classifiers and a multi-label image classifier. The attention adaptor extracts specified features respectively from the deep layer and shallow layer of backbone, then reassigns the weights to different channels. Meanwhile, the domain classifier tries to eliminate domain bias from global to local by learning domain labels with adversarial loss \(\mathcal{L}_{dc}\), the image classifier enforces image-level class-wise alignment on top-level features by learning multi-label recognition with classification loss \(\mathcal{L}_{ic}\). All these techniques progressively implement image-level class-agnostic and class-wise alignment.

I2ITM is an instance-to-image transformation recognition module similar to the shared RPN and detection head, but only enforces image-level predictions for target domain features and supervises the category predictions with a binary cross-entropy loss \(\mathcal{L}_{cls}\). 

The total loss \(\mathcal{L}\) for training our proposed SA3 is summarized as follows:

$$
\mathcal{L}=\underbrace{\lambda_{dc}\mathcal{L}_{dc}+\lambda_{ic}\mathcal{L}_{ic}}_{\text{image-level align}}+\underbrace{\mathcal{L}_{rpn}+\mathcal{L}_{det}+\lambda_{cls}\mathcal{L}_{cls}}_{\text{instance-level align}} \eqno{(1)}
$$
\noindent
where \(\mathcal{L}_{rpn}\) and \(\mathcal{L}_{det}\) are the losses of learning RPN and prediction branch for solely source domain images.

Our framework enables better alignment of crucial regions and important instances across domains. Consequently, the attention-based detection backbone produces more accurate activations on objects of interest in both domains, leading to better adaptive detection performance.

\subsection{Attention-based Global Image-level Alignment}

\begin{figure}
    \centering
    \includegraphics[width=1\linewidth]{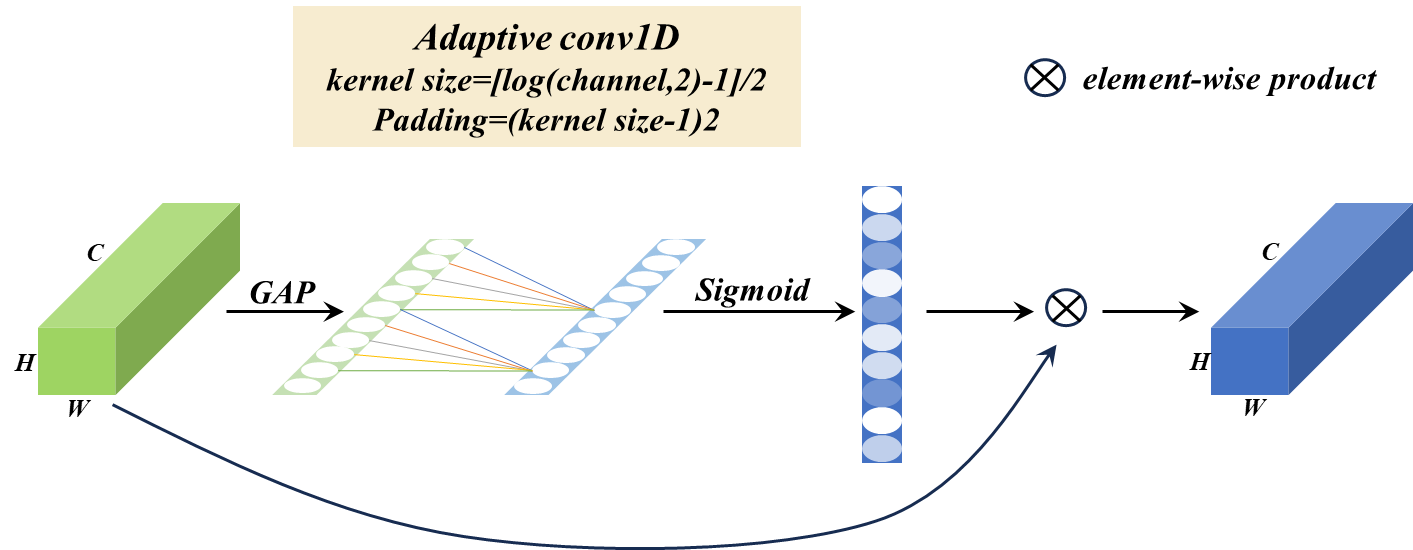}
    \caption{The architecture of Cross-channel Interaction Strategy.}
    \label{fig:333}
\end{figure}

To implement image-level alignment, we make two changes to the backbone:\textbf{ 1)}Feature maps from the top layer of the backbone network have high-level semantics, and we creatively introduce an image-level multi-head classifier combined with an attention mechanism that can adaptively adjust the weights based on the feature contribution values. To achieve this, we propose a better adaptive cross-channel interaction strategy (CIS) for salient features from the backbone network, as shown in \Cref{fig:333}. \textbf{2)}At the same time, the domain labels of all the images are freely available whether from the source or target domains, using domain labels, we build the adversarial training to mitigate the discrepancy between global features from the source and target domain, performing the attention-based image-level class-agnostic alignment.

CIS endeavors to adaptively determine the coverage of local 
cross-channel interaction and adjust the\textit{ \(k\)}-channels’ weights according to the attention scores cucullated with fast 1D convolution of size\textit{ \(k\)}. Inspired by SENet \cite{b15}, we inherited the basic principle of cross-channel interaction but it’s inefficient and unnecessary to capture dependencies across all channels, thus we restructure a novel local dependency strategy and implement the mapping in non-linear function between the coverage of interaction (kernel size of 1D convolution) and the channel dimension\textit{ \(C\)}. Since the channel dimension \textit{\(C\)} is usually set to a power of 2, we extend the linear function $\phi(k)=\gamma*k+b$ to the non-linear one, i.e.,

    $$
    C=\phi(k)=2^{(\gamma*k+b)}  \eqno{(2)}
    $$

Then, given channel dimension \textit{\(C\)}, kernel size k can be adaptively determined by 
    
    $$
    k=\psi(C)=\left|\frac{log_2(C)}\gamma-\frac b\gamma\right|_{odd}  \eqno{(3)}
    $$

\noindent
where \(|t|_{odd}\) indicates the nearest odd number of \textit{\(t\)}. In this paper, we set \textit{\(\gamma\)} and\textit{ \(b\)} to 2 and 1 throughout all the experiments, respectively. Giving the aggregated features $y\in\mathbb{R}^{C}$ without dimensionality reduction, channel attention can be learned by

    $$
    \omega=\sigma(\mathcal{W}y)  \eqno{(4)}
    $$
    
\noindent
where \(\sigma(\cdot)\) is a sigmoid function and \textit{\(\mathcal{W}\)} is an attention matrix. With the coverage of channel interaction \textit{\(k\)}, we could set the fast 1D convolution and employ the bank matrix \textit{\(\mathcal{W}_k\)} to learn channel attention only considering local interaction between \textit{\(y_i\)} and its \textit{\(k\)} neighbors. Let the output of the backbone be $\chi\in\mathbb{R}^{W\times H\times C}$, accordingly the weights of channels in this block can be computed as:

    $$
    \omega=\sigma\left(C1D_{\{k\}}(g(\chi))\right) \eqno{(5)}
    $$

\noindent
where $g(\mathcal{X})=\frac{1}{WH}\sum_{i=1,j=1}^{W,H}\chi_{ij}$ is channel-wise global average pooling. Thus, through this attention block, we can realize cross-channel interactions over long-range scales in deep feature layers with high channel dimensions.

Combining with the CIS, we construct an attention-based category-wared alignment module by attaching an image-level multi-label image classifier. Using the available image-level annotations, we can easily compute the binary cross-entropy loss and optimize the parameters of the attention network and classifier under the classification supervision. During training, the prototypes learned from the image classifier pull the features from the corresponding classes towards themselves, regardless of the underlying domain, enabling attention-based image-level alignment. Also, we attach two domain classifiers to the backbone to align deep and shallow features simultaneously, similar to SWDA \cite{b16}, but the attention module only works on the deep features input to the domain classifiers. 

Hence, the loss of AIAM for realizing the domain discrepancy image-level alignment from global to local can be defined as:

    $$
    L_{\mathrm{AGIA}}=\lambda_{dc}L_{dc}+\lambda_{ic}L_{ic}  \eqno{(6)}
    $$

\noindent
where \textit{\(\lambda_{dc}\) }and \textit{\(\lambda_{ic}\)} are the hyper-parameters used to control the weighting of corresponding losses.

\subsection{Instance-to-image Level Transformation Recognition}

\begin{figure}
    \centering
    \includegraphics[width=1\linewidth]{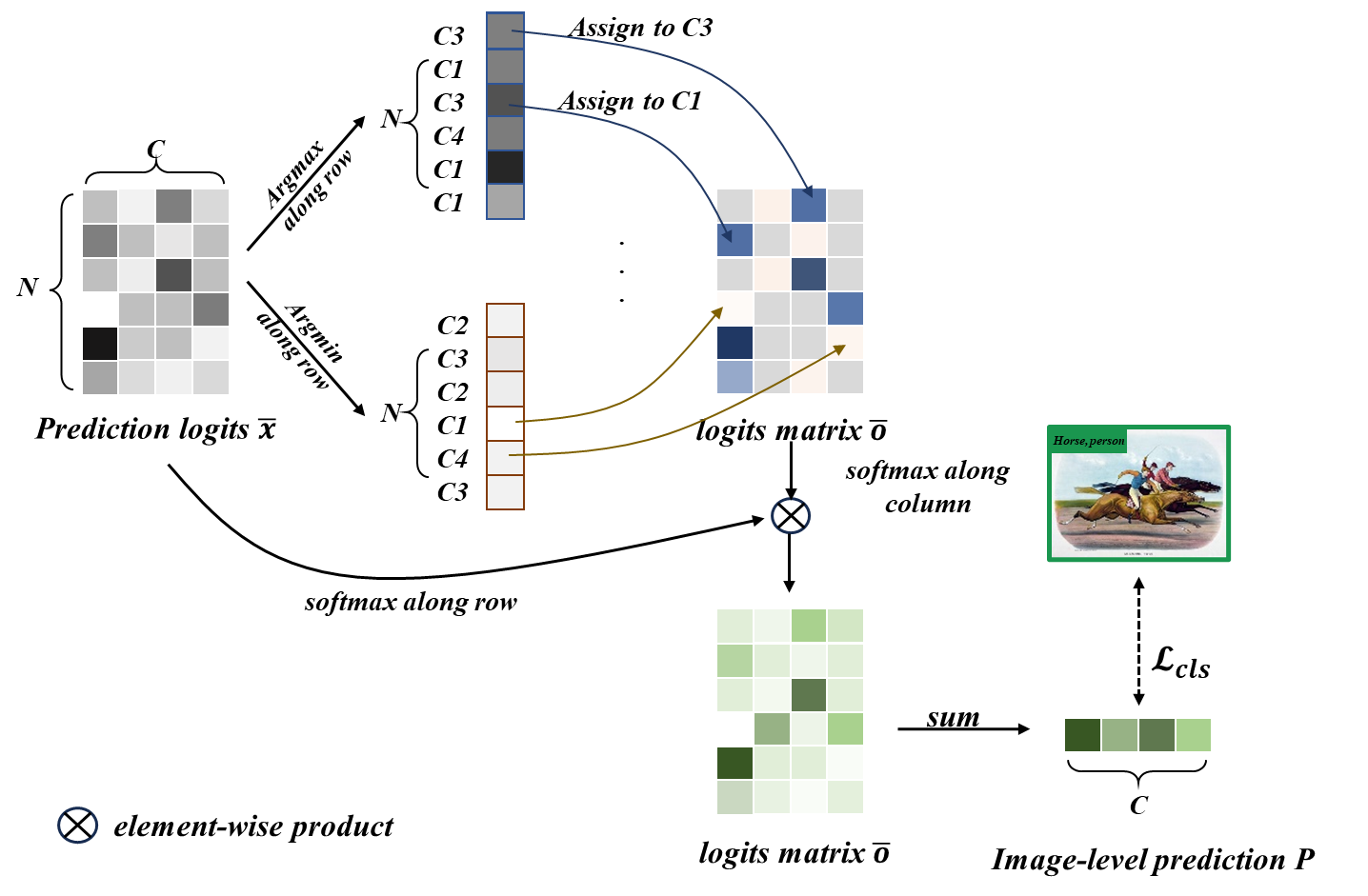}
    \caption{Instance-to-image Level Transformation Module.}
    \label{fig:444}
\end{figure}

After global image-level alignment, our work focuses on how to further enforce instance-level alignment. We propose a new instance alignment method I2ITM, which aggregates multiple instance-level outputs from RPN and detection head into image-level multi-label prediction to achieve weak supervision, as depicted in \Cref{fig:444}.

I2ITM aggregates the objectness logits \(o\in\mathbb{R}^N\) with the classification logits $x\in\mathbb{R}^{N\times\mathcal{C}}$(\(\mathcal{C}\) is the total number of object classes), where the former is obtained from the RPN of the original pipeline, and latter is obtained from the detection head, constituting the instance-level multi-label prediction logits \(\bar{x}\in\mathbb{R}^{N\times\mathcal{C}}\). We aim to transform it into an image-level classification prediction, in which we introduce a novel self-aware transformation method, specifically, we first compute a new logits matrix \(\bar{o}\in\mathbb{R}^{N\times\mathcal{C}}\) based on \(\bar{x}\), where we assign the \textit{\(n\)}-th objectness \(o_n\) to a specific object class along the row according to \(x_n\in\mathbb{R}^{C}\): if the index for the largest-value entry of \(x_n\) is\textit{ \(i\)} we assign on to the \(i\)-th entry, if the index for the smallest-value entry of \(x_n\) is \textit{\(j\)} we assign its negative value to the\textit{ \(j\)}-th entry, all the other entries assigned 0 objectness. In this way, we obtain a class-specific objectness matrix \(\bar{o}\in\mathbb{R}^{N\times\mathcal{C}}\).

Then, we apply the softmax function on \(\bar{x}\) along object classes and \(\bar{o}\) along proposals respectively, to extract the weighted proposals' scores of \(c\) object classes (softmax along row in \Cref{fig:444}) and proposals (softmax along column). Finally, we can obtain the image-level predictions by making element-wise product operations between softmax \(\bar{x}\) and softmax \(\bar{o}\). Formally, the image-level prediction aggregation can be given as:
    $$
    P_c=\sum_{n=1}^N(\sigma^{row}(\bar{x}_{n,c})\odot\sigma^{col}(\bar{o}_{n,c}))  \eqno{(7)}
    $$

\noindent
where \textit{\(\mathcal{P}_c\)} is the predicted probability for class \(c\) (indicating whether current image contains objects of class \textit{\(c\)}), \(\sigma^{row}(\cdot)\) and \(\sigma^{col}(\cdot)\) are softmax operations along row and column respectively, and \(\odot\) is the element-wise product operation.

\begin{table*}[htbp]
 \renewcommand{\arraystretch}{1.1}
 \setlength{\tabcolsep}{3.5pt}
    \centering
    \caption{The results of SA3 on the Clipart-test for VOC→ Clipart1k adaptation.}
    \begin{tabular}{l | ccccc ccccc ccccc ccccc |c } 
\hline
 Method & aero   & bike    & bird           & boat   & bottle  & bus             & car    & cat     & chair           & cow    & table   & dog             & hrs    & m-bike  & prsn            & plnt   & sheep   & sofa            & train  & tv      & mAP  \\ 
\hline
Source  & 22.7                & 55.0                & 23.0                & 30.9                & 43.3                & 49.8                & 29.6                & 19.3                & 42.0                & 13.0                & 32.0                & 10.9                & 24.5                & 59.9                & 40.4                & 50.0                & 18.2                & 28.3                & 43.9                & 42.8                & 34.0                 \\ 
\hline
CRDA    & 28.7                & 55.3                & 31.8                & 26.0                & 40.1                & 63.6                & 36.6                & 9.4                 & 38.7                & 49.3                & 17.6                & 14.1                & 33.3                & 74.3                & 61.3                & 46.3                & 22.3                & 24.3                & 49.1                & 44.3                & 38.3                 \\
HTCN    & 33.6                & 58.9                & 34.0                & 23.4                & 45.6                & 57.0                & 39.8                & 12.0                & 39.7                & 51.3                & 21.1                & 20.1                & 39.1                & 72.8                & 63                  & 43.1                & 19.3                & 30.1                & 50.2                & 51.8                & 40.3                 \\
DM      & 25.8                & 63.2                & 24.5                & 42.4                & 47.9                & 43.1                & 37.5                & 9.1                 & 47.0                & 46.7                & 26.8                & 24.9                & 48.1                & 78.7                & 63                  & 45.0                & 21.3                & 36.1                & 52.3                & 53.4                & 41.8                 \\
UMT\cite{b3}  & 39.6                & 59.1                & 32.4                & 35.0                & 45.1                & 61.9                & 48.4                & 7.5                 & 46.0                & 67.6                & 21.4                & 29.5                & 48.2                & 75.9                & 70.5                & 56.7                & 25.9                & 28.9                & 39.4                & 43.6                & 44.1                 \\
PLGE\cite{b4} & 43.4                & 52.5                & 29.4                & 40.1                & 30.4                & 71.9                & 54.9                & 3.6                 & 52.4                & 73.8                & \textbf{53.5}\par{} & 24.0                & 54.8                & \textbf{89.1}\par{} & 65.1                & 40.5                & 32.3                & 33.8                & 45.4                & \textbf{61.0}\par{} & 47.6                 \\
AT\cite{b5}   & 33.8                & 60.9                & 38.6                & 49.4                & 52.4                & 53.9                & \textbf{56.7}\par{} & 7.5                 & 52.8                & 63.5                & 34.0                & 25.0                & \textbf{62.2}\par{} & 72.1                & \textbf{77.2}\par{} & 57.7                & 27.2                & \textbf{52.0}\par{} & 55.7                & 54.1                & 49.3                 \\
H2FA[2] & 38.5                & \textbf{70.6}\par{} & \textbf{70.6}\par{} & 47.4                & 59.6                & \textbf{83.5}\par{} & 47.0                & 29.3                & 51.5                & 76.3                & 44.4                & \textbf{48.1}\par{} & 47.3                & 79.2                & 75.7                & 54.4                & \textbf{53.9}\par{} & 32.0                & 56.6                & 51.1                & 55.3                 \\
Ours    & \textbf{47.5}\par{} & 60.1                & 43.9                & \textbf{55.3}\par{} & \textbf{60.2}\par{} & 70.1                & 55.6                & \textbf{29.3}\par{} & \textbf{55.7}\par{} & \textbf{84.9}\par{} & 40.5                & 47.1                & 57.3                & 74.0                & 76.3                & \textbf{60.0}\par{} & 53.8                & 42.4                & \textbf{62.0}\par{} & 44.9                & \textbf{56.0}\par{}  \\ 
\hline
Oracle  & 55.2                & 78.3                & 51.1                & 58.1                & 60.7                & 58.4                & 61.5                & 27.3                & 60.9                & 71.7                & 60.5                & 40.7                & 56.9                & 82.5                & 82.8                & 65.9                & 49.2                & 46.1                & 59.7                & 58.1                & 59.3                 \\
\hline
\end{tabular}
    \label{tab:111}
\end{table*}
\begin{table*}[htbp]
 \renewcommand{\arraystretch}{1.1}
 \setlength{\tabcolsep}{3.3pt}
    \centering
    \caption{The results of SA3 on the Clipart-all for VOC→Clipart1k adaptation.}
    \begin{tabular}{l | ccccc ccccc ccccc ccccc | c } 
\hline
 Method & aero   & bike    & bird           & boat   & bottle  & bus             & car    & cat     & chair           & cow    & table   & dog             & hrs    & m-bike  & prsn            & plnt   & sheep   & sofa            & train  & tv      & mAP  \\ 
\hline
        Source    & 32.0          & 44.4          & 28.7          & 28.8          & 40.8          & 57.6          & 35.3          & 16.8          & 41.2          & 23.8          & 27.3          & 15.9          & 22.5          & 77.3           & 36.5          & 46.4          & 9.1           & 27.8          & 46.9          & 47.1          & 35.3           \\ 
\hline
PCL       & 3.4           & 10.6          & 2.3           & 1.7           & 5.2           & 3.4           & 23.3          & 1.2           & 5.6           & 0.4           & 7.8           & 3.7           & 5.6           & 0.3            & 24.5          & 19.7          & 11.9          & 3.6           & 9.2           & 25.4          & 8.4            \\
SWDA\cite{b16}  & 26.2          & 48.5          & 32.6          & 33.7          & 38.5          & 54.3          & 37.1          & 18.6          & 34.8          & 58.3          & 17.0          & 12.5          & 33.8          & 65.5           & 61.6          & 52.0          & 9.3           & 24.9          & 54.1          & 49.1          & 38.1           \\
DBGL  & 28.5          & 52.3          & 34.3          & 32.8          & 38.6          & 66.4          & 38.2          & 25.3          & 39.9          & 47.4          & 23.9          & 17.9          & 38.9          & 78.3           & 61.2          & 51.7          & 26.2          & 28.9          & 56.8          & 44.5          & 41.6           \\
IIOD  & 41.5          & 52.7          & 34.5          & 28.1          & 43.7          & 58.5          & 41.8          & 15.3          & 40.1          & 54.4          & 26.7          & 28.5          & 37.7          & 75.4           & 63.7          & 48.7          & 16.5          & 30.8          & 54.5          & 48.7          & 42.1           \\
ICCM\cite{b1}   & 39.8          & 66.7          & 37.2          & 42.5          & 43.3          & 48.1          & 48.1          & 21.3          & 46.5          & 73.0          & 29.0          & 29.8          & 57.3          & 78.6           & 67.8          & 48.7          & 46.3          & 19.3          & 42.8          & 48.5          & 46.7           \\
DT+PL & 50.1          & \textbf{75.0} & 37.0          & 38.7          & 58.1          & 83.4          & 50.1          & 38.0          & 55.2          & 67.3          & 51.1          & 34.8          & 49.8          & 89.9           & 60.2          & 63.4          & 28.8          & 42.4          & 62.6          & 70.9          & 55.3           \\
H2FA\cite{b2}   & 58.1          & 73.0          & \textbf{56.8} & \textbf{50.4} & 61.2          & 98.6          & \textbf{69.5} & 57.8          & \textbf{66.4} & 77.1          & 56.1          & \textbf{84.1} & \textbf{64.3} & 100.0          & 78.1          & \textbf{78.2} & 43.5          & 65.4          & 77.3          & 79.7          & 69.8           \\
Ours      & \textbf{60.0} & 69.5          & 47.1          & 47.9          & \textbf{69.3} & \textbf{99.6} & 67.9          & \textbf{60.1} & 66.0          & \textbf{81.6} & \textbf{59.1} & 77.1          & 60.4          & \textbf{100.0} & \textbf{79.6} & 77.3          & \textbf{49.7} & \textbf{73.8} & \textbf{79.1} & \textbf{84.1} & \textbf{70.5}  \\
\hline
    \end{tabular}
    \label{tab:222}
\end{table*}

With image-level multi-label predictions and one-hot encoding labels, we employ a binary cross-entropy loss \textit{\(\mathcal{L}_{cls}\)} for optimization, mitigating the domain discrepancy of instance-level features through weak supervision training.

\section{Experiments}

\subsection{Datasets}

To illustrate the application of our framework, we select four public datasets to conduct our experiments, including PASCAL VOC2007, PASCAL VOC2012, Clipart1k, and Comin2k. Following H2FA \cite{b2}, we set the public benchmark VOC (PASCAL VOC 2007 and PASCAL VOC 2012) as the source domain in whole experiments, while the other datasets, formed by unrealistic images with a distinct cartoon style, are treated as target domains respectively. In the cross-domain weakly supervised task, the source domain provides instance-level annotations for training, while the target domain only provides image-level annotations.

\textbf{PASCAL VOC. }VOC contains PASCAL VOC 2007 and PASCAL VOC 2012, whose trainval sets are used as the source domain training data in the experiments. There are 16551 real-world images of 20 categories in the trainval set of VOC, each attaching instance-level labeling information.
\textbf{Clipart1k. }Clipart1k has a train split and a test split, each split consists of 500 images from the same 20 object categories as the PASCAL VOC. Following prior works \cite{b3,b4,b5}, we train domain adaptation only using the training set as unlabeled target data, and the testing set is held out for evaluation (termed as Clipart-test). Besides, like \cite{b1,b16}, for Clipart1k all its 1000 images are also used for training/evaluation (termed as Clipart-all).
\textbf{Comic2k.} Comic2k has the same split ratio as Clipart1k, each split contains 1000 images for training/testing, a total of 2000 images from 6 classes, shared with the same classes in the PASCAL VOC dataset. Considering
Considering above datasets, we set two scenarios for cross domian task, Real-to-Clipart and Real-to-Comic, including three benchmarks.

\subsection{Implementation Details}

Following H2FA \cite{b2} and AT \cite{b5}, we employ a two-stage detector Faster R-CNN \cite{b11} with RoIAlign as the base framework in our proposed method SA3, implement and evaluate it using Detectron2. ImageNet pre-trained ResNet-101  is utilized as our network backbone in all experiments.

The experiments are all conducted using a single Nvidia GPU V100 trained at a mini-batch size of 8 (4 images per domain), the initial learning rate is set to 0.005, the number of training iterations is set as 24k, and the learning rate is multiplied by 0.1 to continue training iteratively at 16k and 21.5k iterations. We empirically set the loss weights \textit{\(\lambda_{dc}\)}, \textit{\(\lambda_{ic}\)}, and \textit{\(\lambda_{cls}\)} to 1, 0.1, and 1. Other hyper-parameters are the default setups in Detectron2, and we do not tune them ad hoc.

\subsection{Results and Comparisons}

In this section, we report the performance of our Attention-based Adaptive Alignment and other state-of-the-art approaches. Additionally, we report the source-only baseline trained on the source domain as the lower bound benchmark. On the other hand, we report the oracle result obtained by training the model with the same target images but annotated with ground truth annotations, which can be viewed as the reference for the upper bound adaptation performance. We report the average precision (AP) of each category as well as mean Average Precision (mAP ) over all categories for object detection on the Clipart1k and Comic2k datasets.

\begin{table}
    \renewcommand{\arraystretch}{1.2}  
 \setlength{\tabcolsep}{4.5pt}   
    \centering
    \caption{The results of SA3 on the Comic2k test set for VOC→Comic2k adaptation.}
    \begin{tabular}{l | ccc ccc | c } 
         \hline
Method    & bicycle & bird  & car  & cat   & dog  & person & mAP    \\ 
\hline
Source    & 49.4     & 12.6   & 21.2  & 11.6   & 19.4  & 37      & 25.2    \\
SWDA\cite{b16}  & 30.3     & 19.6   & 28.8  & 15.2   & 24.9  & 46.9    & 27.6    \\
DBGL      & 35.6     & 20.3   & 33.9  & 16.4   & 26.6  & 45.3    & 29.7    \\
MCAR      & 47.9    & 20.5   & 37.4  & 20.6   & 24.5  & 50.2    & 33.5    \\
DT+PL & 55.2    & 18.5   & 38.2  & 22.9   & 34.1  & 54.5    & 37.2    \\
PLGE\cite{b4}   & 55.0      & 21.2   & 40.0    & 35.1   & 37.9  & 60.9    & 41.7    \\
H2FA\cite{b2}   & 55.3    & 26.6   & \textbf{45.9} & 38.1   & \textbf{45.6} & 66.8    & 46.4    \\
Ours      & \textbf{56.9}    & \textbf{29.0}  & 43.8  & \textbf{42.6}  & 43.8  & \textbf{67.1}   & \textbf{47.2}   \\ 
\hline
Oracle             & 61.9     & 38.9   & 50.8          & 48.9   & 45.2          & 76.6    & 53.7    \\
\hline
    \end{tabular}
    \label{tab:333}
\end{table}

\begin{figure*}
	\centering
	\begin{subfigure}{0.23\linewidth}
		\centering
		\includegraphics[width=0.9\linewidth]{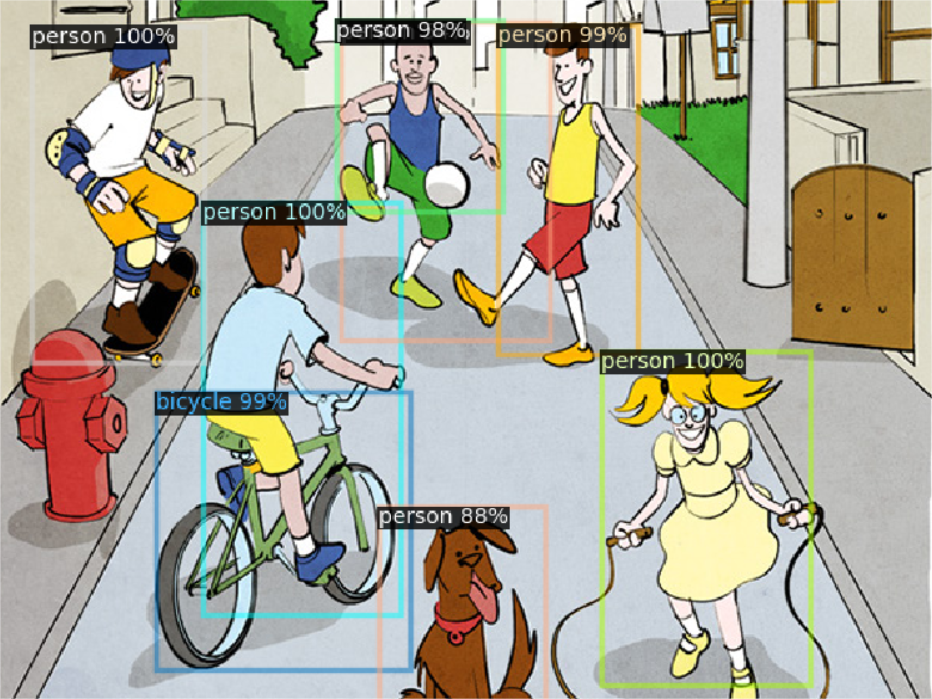}
		\label{51}
	\end{subfigure}
	\begin{subfigure}{0.23\linewidth}
		\centering
		\includegraphics[width=0.9\linewidth]{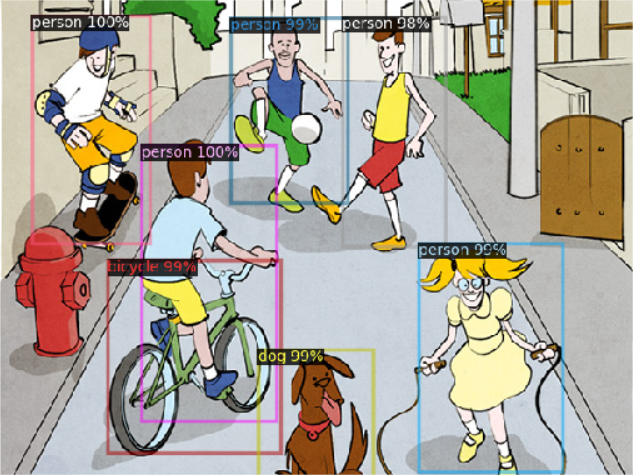}
		\label{52}
	\end{subfigure}
        \begin{subfigure}{0.23\linewidth}
		\centering
		\includegraphics[width=0.9\linewidth]{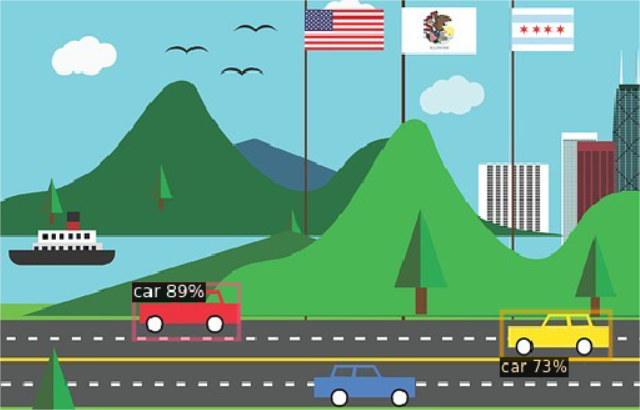}
		\label{53}
	\end{subfigure}
        \begin{subfigure}{0.23\linewidth}
		\centering
		\includegraphics[width=0.9\linewidth]{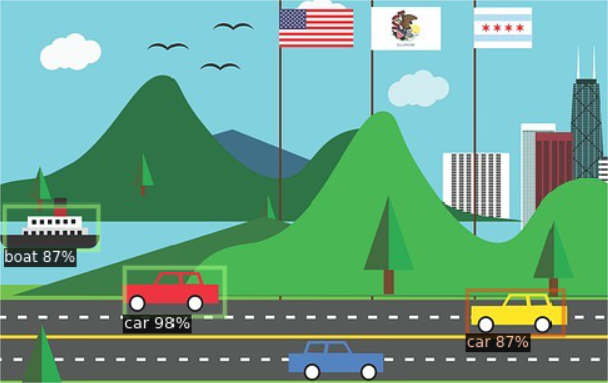}
		\label{54}  
	\end{subfigure}
	
	\begin{subfigure}[b]{0.23\linewidth}
		\centering
		\includegraphics[width=0.9\linewidth]{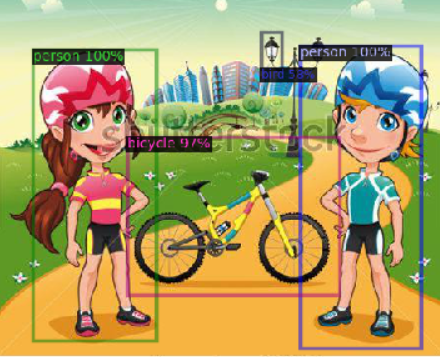}
		\caption{H2FA}
		\label{55}
	\end{subfigure}
	\begin{subfigure}[b]{0.23\linewidth}
		\centering
		\includegraphics[width=0.9\linewidth]{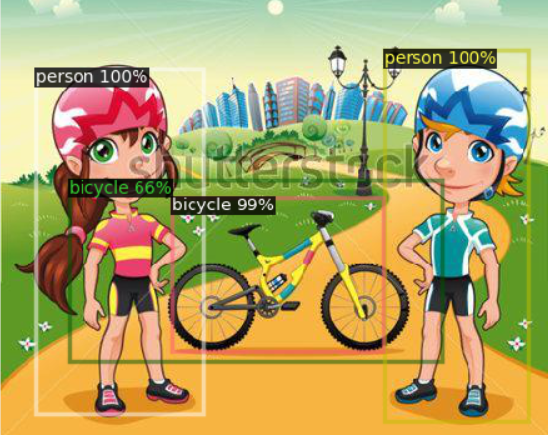}
		\caption{SA3 R-CNN}
		\label{56}
	\end{subfigure}
        \begin{subfigure}[b]{0.23\linewidth}
		\centering
		\includegraphics[width=0.9\linewidth]{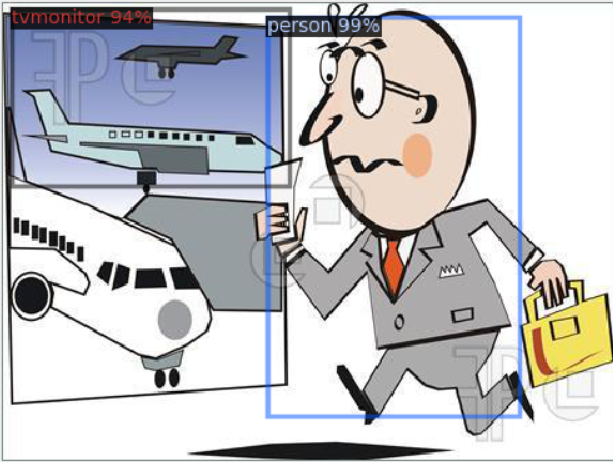}
		\caption{H2FA}
		\label{57}
	\end{subfigure}
        \begin{subfigure}[b]{0.23\linewidth}
		\centering
		\includegraphics[width=0.9\linewidth]{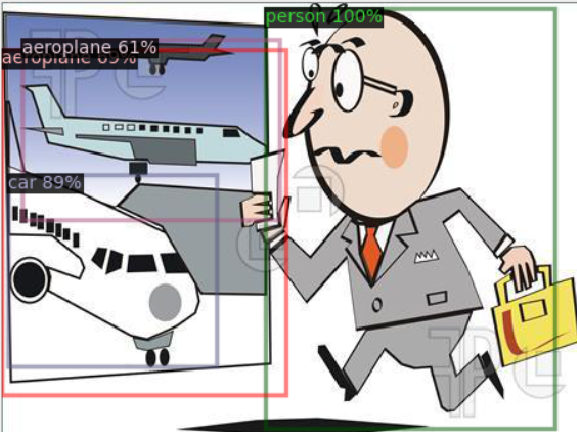}
		\caption{SA3 R-CNN}
		\label{58}
	\end{subfigure}
    \caption{Illustration of the detection results on the target domain of Clipart1k-test.}
    \label{fig:555}
\end{figure*}
\begin{figure}
	\centering
	\begin{subfigure}{0.49\linewidth}
		\centering
		\includegraphics[width=1\linewidth]{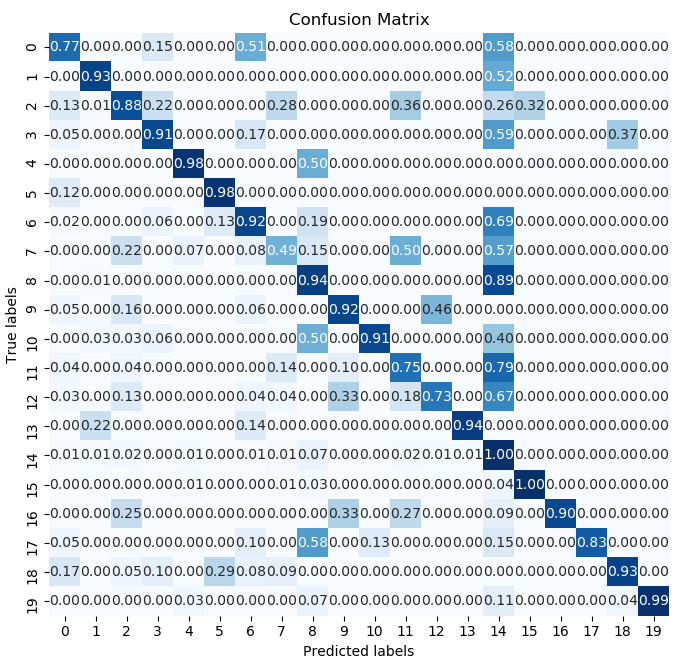}
            \caption{H2FA}
		\label{61}
	\end{subfigure}
        \begin{subfigure}{0.49\linewidth}
		\centering
		\includegraphics[width=0.99\linewidth]{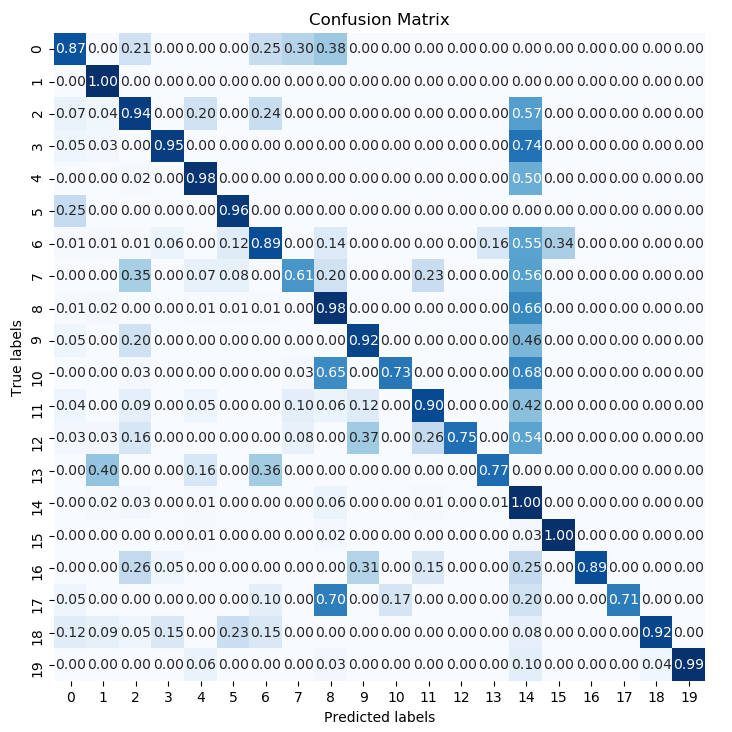}
            \caption{Ours}
		\label{63}
	\end{subfigure}
    \caption{Global relation matrices of Pascal VOC→Clipart1k-test.}
    \label{fig:666}
\end{figure}

\textbf{1) Scenario 1: Real-to-Clipart}

As is shown in Table \ref{tab:111}, we can observe the mean AP performance on Clipart-test, SA3 brings significant improvement (from 34.0\% to 56.0\% mAP) over the source-only model. Also, compared with the previous works, SA3 achieves state-of-the-art performance and outperforms the recently superior method H2FA by 0.7\% in terms of mAP. 
Several comparative detection samples are illustrated in \Cref{fig:555}, the model's robustness to domain shift is further evident in these results. The confusion matrix, as shown in \Cref{fig:666}, reveals that SA3 effectively reduces false positives and increases true positives, leading to improved detection accuracy. 

From the following Table \ref{tab:222}, our SA3 shows improved performance compared to H2FA and reaches a mean AP of 70.5\%, which outperforms the result of 35.3\% mAP from the source-only baseline. We note that, the performance in detecting bus and m-bike classes is remarkably superior (99.6\% and 100.0\% mAP), with excellent domain adaptation.

\textbf{2) Scenario 2: Real-to-Comic}

Scenario 2 corresponds to the results of Table \ref{tab:333}, the Comic2k benchmark is a particularly challenging benchmark, where the source-only model only achieves 25.2\% mAP. From the results obtained so far, it clearly seems that SA3 exceeds all the previous methods at 47.2\% mAP and surpasses the second-place model, H2FA, by 0.8\%. This significant improvement in performance over previous methods greatly mitigates the gap to the Oracle model, which achieves a mAP of 53.7\%. These results demonstrate the superiority of the SA3 model on the Comic2k benchmark.

\subsection{Ablation study}

In the field of domain adaption, ablation studies are essential for investigating the impact of major components in a model. In this context, we further conduct ablation studies on three benchmarks. We use Baseline,  ECA (Efficient Channel Attention), and CIS to indicate the baseline model, model with ECA, and model with the proposed Cross-channel Interaction Strategy (CIS) for brevity. The baseline adopts only common practices and AIAM without any attention module for aligning the image-level features.

For our analysis, we consider the mAP scores as a measure of performance, which stands for mean Average Precision and a common metric used in object detection tasks. We compare the mAP scores of the three models across all datasets to investigate the superiority of the proposed CIS and present the qualitative studies in Table \ref{tab:444}.

On\textit{ clipart-test}, our CIS achieves a mAP score of 56.0\%, which is higher than both the baseline of 55.3\% and ECA of 54.4\% mAP. However, on \textit{clipart-all}, CIS fell slightly short of ECA's performance, with an mAP of 70.5 versus ECA's 74.3\% mAP. While these results demonstrate the effectiveness of CIS, we still believe that there remains potential room for further optimization. Notably, on the challenging \textit{Comic2k} benchmark, CIS exhibited a competitive mAP score of 47.2\%, surpassing both the baseline and ECA. 

In general, while the absolute mAP scores vary across datasets, the relative performance trends remain consistent: our CIS outperforms the baseline and ECA model. Although CIS falls short of ECA in terms of mAP scores on certain datasets, it is important to note that CIS still demonstrate superiority through other metrics or in specific use cases.

\begin{table}
   \renewcommand{\arraystretch}{1.3}  
   \setlength{\tabcolsep}{12pt}   
    \centering
    \caption{Effectiveness of different attention blocks in SA3 R-CNN, where mean AP performance (\%) over all classes is reported.}
    \begin{tabular}{l | ccc  } 
         \hline
~         & Clipart-test                                   & Clipart-all                                    & Comic2k                                         \\ 
\hline
Baseline& 55.3                                           & 69.8                                           & 46.4                                            \\
ECA& 54.4\textcolor{c2}{(-0.9)}& 74.3\textcolor{c1}{(+4.5)}& 46.7\textcolor{c1}{(+0.3)}\\
CIS(Ours)& 56.0\textcolor{c1}{(+0.7)}& 70.5\textcolor{c1}{(+0.7)}& 47.2\textcolor{c1}{(+0.8)}\\
\hline
    \end{tabular}
    \label{tab:444}
\end{table}

\section{Conclusion}
 
This paper proposes the framework of the Self-Aware Adaptive Alignment for Cross-Domain Object Detection. Through the validation on widely used benchmarks, it's proven that our proposed method holds great potential for enhancing object detection in cross-domain settings, and it paves the way for many exciting applications involving intelligent traffic monitor systems and virtual reality in the future.






\end{document}